\pdfoutput=1
\documentclass[11pt]{article}

\usepackage[final]{coling}

\usepackage{times}
\usepackage{latexsym}
\usepackage{booktabs}
\usepackage{multirow}
\usepackage{float}
\usepackage{stfloats}
\usepackage{subfig}

\usepackage[T1]{fontenc}
\usepackage[utf8]{inputenc}

\usepackage{microtype}

\usepackage{inconsolata}

\usepackage{graphicx}

\usepackage{amsmath,amsfonts}
\usepackage{algorithmic}
\usepackage{algorithm}

\usepackage{hyperref}

\title{Dynamic Graph Neural Ordinary Differential Equation Network for Multi-modal Emotion Recognition in Conversation}

\author{Yuntao Shou$^{1,2}$, Tao Meng$^{3,\dagger}$, Wei Ai$^{3}$, Keqin Li$^{4}$ \\
	$^1$School of Computer Science and Technology, Xi’an Jiaotong University, Xi’an, 710049, China
	 \\
	$^2$Ministry of Education Key Laboratory of Intelligent Networks and Network Security,\\
	Xi’an Jiaotong University, Xi’an, 710049, China\\
	$^3$College of Computer and Mathematics, Central South University of Forestry and Technology, \\ Changsha, Hunan, 410004, China \\
	$^4$Department of Computer Science, State University of New
	York, New Paltz, New York 12561, USA \\
\textbf{$^\dagger$Corresponding Author: mengtao@hnu.edu.cn} 
}

\begin{document}
\maketitle
\begin{abstract}
Multimodal emotion recognition in conversation (MERC) refers to identifying and classifying human emotional states by combining data from multiple different modalities (e.g., audio, images, text, video, etc.). Most existing multimodal emotion recognition methods use GCN to improve performance, but existing GCN methods are prone to overfitting and cannot capture the temporal dependency of the speaker's emotions. To address the above problems, we propose a Dynamic Graph Neural Ordinary Differential Equation Network (DGODE) for MERC, which combines the dynamic changes of emotions to capture the temporal dependency of speakers' emotions, and effectively alleviates the overfitting problem of GCNs. Technically, the key idea of DGODE is to utilize an adaptive mixhop mechanism to improve the generalization ability of GCNs and use the graph ODE evolution network to characterize the continuous dynamics of node representations over time and capture temporal dependencies. Extensive experiments on two publicly available multimodal emotion recognition datasets demonstrate that the proposed DGODE model has superior performance compared to various baselines. Furthermore, the proposed DGODE can also alleviate the over-smoothing problem, thereby enabling the construction of a deep GCN network.
\end{abstract}

\section{Introduction}
\label{sec:intro}

\begin{figure}
	\centering
	\includegraphics[width=0.9\linewidth]{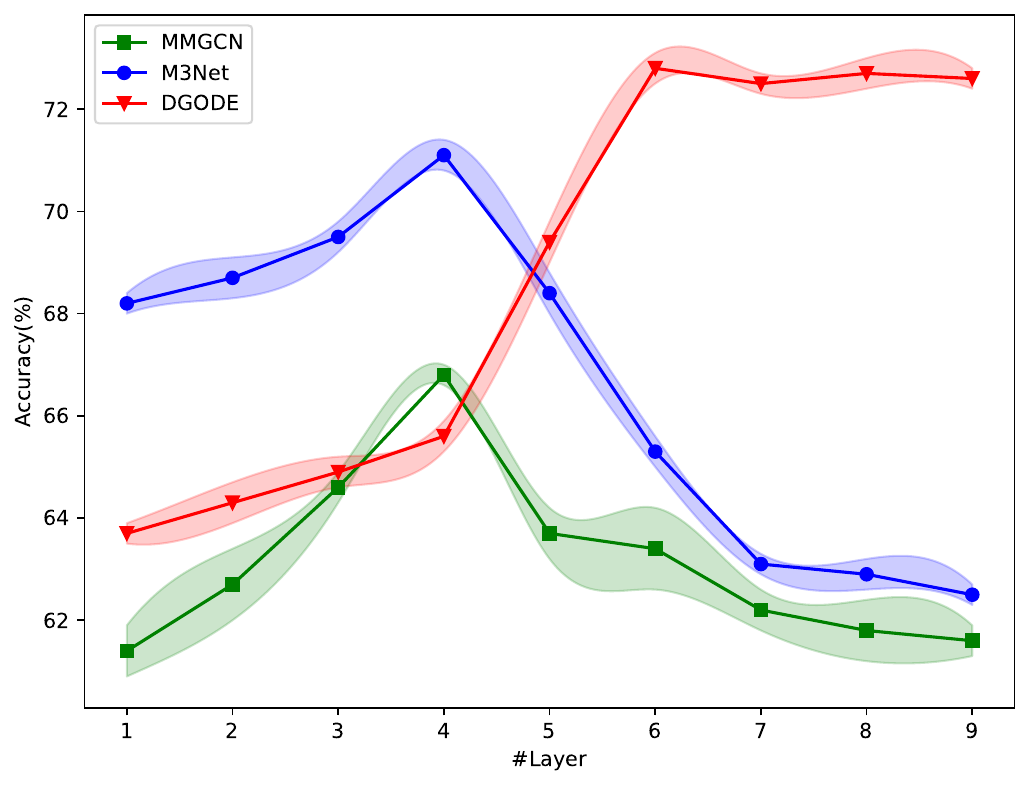}
	\caption{Performance comparison of different methods for different numbers of GCN layers on the IEMOCAP dataset.}
	\label{fig:accuracyvslayer}
	\vspace{-5mm}
\end{figure}

Multimodal Emotion Recognition in Conversation (MERC) technology significantly improves the accuracy and wide application of emotion recognition by integrating data from multiple modalities (e.g., audio, image, text, and video) \cite{shou2022conversational, shou2025masked, shou2022object, shou2023comprehensive, shou2024adversarial, meng2024deep, shou2023adversarial, ai2023gcn, ai2024gcn, meng2024multi, shou2024contrastive, shou2024spegcl, shou2024efficient}. MERC can not only improve the intelligence of human-computer interaction, but also bring important improvements in practical scenarios (e.g., health monitoring, education, entertainment, and security) \cite{ying2021prediction, shou2023graph, shou2023czl, meng2024masked, shou2024revisiting, ai2024edge, zhang2024multi, ai2023two, meng2024revisiting, ai2024mcsff}.

Many existing studies improve the performance of MERC by using graph convolutional neural networks (GCNs) \cite{yin2023omg, yin2023coco, yin2023messages, yin2022deal, yin2022dynamic, yin2024dynamic, yin2024continuous, yinsport} to effectively model the conversational relations between speakers. However, as shown in Fig. 1, we find that existing GCNs only contain 4 layers of GCN (e.g., MMGCN \cite{hu2021mmgcn} and M3Net \cite{chen2023multivariate}), while the performance decreases significantly as the number of layers increases. The reason for the performance degradation may be attributed to the fact that information aggregation in vanilla GCNs is just simple message smoothing within the neighborhood, causing neighboring nodes to converge to the same value as the number of layers is stacked. Therefore, it is necessary to improve the stability of information diffusion on the graph and alleviate the problem of over-smoothing of nodes. Furthermore, MERC usually relies on dynamic temporal information, while traditional GCNs can only process static images. How to model the temporal features in emotion recognition tasks as a continuous dynamic process so that the network can capture and model the complex patterns of emotion changing over time remains a challenge. Therefore, this paper aims to propose a Dynamic Graph Neural Ordinary Differential Equation Network (DGODE) to dynamically model the temporal dependency of emotion changes and improve the expressiveness of node features as the number of GCN layers increases \cite{shou2023graphunet, ai2024graph, shou2024low, shou2024graph, ai2024seg, ai2024contrastive}.

Specifically, we introduce the Dynamic Graph Neural Ordinary Differential Equation Network (DGODE) based on the perspective of continuous time. Our DGODE method introduces an adaptive mixhop mechanism to extract node information from different hop count neighbors simultaneously and uses ordinary differential equations to model the temporal dependence of emotion changes. DGODE shows stable performance as the number of GCN layers increases. We use two publicly available multimodal emotion recognition datasets to verify the effectiveness of DGODE.

Overall, the main contributions of this paper are as follows:

\begin{itemize}
	\item We propose the Dynamic Graph Neural Ordinary Differential Equation Network (DGODE), which combines the dynamic changes of emotions to capture the temporal dependency of speakers' emotions.
	
	\item We design an adaptive mixhop mechanism to capture the relationship between distant nodes and combine ODE to capture the temporal dependency of speakers' emotions.
	
	\item Extensive experiments are conducted to demonstrate its superiority in MERC compared to various baselines on IEMOCAP and MELD datasets.
\end{itemize}

\section{Related Work}

\subsection{Multimodal Emotion Recognition in Conversation}

MERC aims to identify and understand the emotional state in a conversation by analyzing data in multiple modalities \cite{zhang2023dualgats}. With the rapid development of social media technology, people are increasingly communicating in a multimodal way. Therefore, how to accurately understand the emotional information in multimodal data has become a key issue \cite{ai2024gcn}.

Initially, researchers used recurrent neural networks (RNNs) to model conversations, which mainly capture emotional information by processing utterances or entire conversations sequentially \cite{poria2017context}. For instance, HiGRU \cite{jiao2019higru} proposed a hierarchical GRU model to capture the information in the conversation, which not only considers the emotional features at the word level, but also extends to the utterance level, thereby generating a conversation representation that contains richer contextual information. Similarly, DialogueRNN \cite{majumder2019dialoguernn} also uses GRU units to capture the emotional dynamics in the conversation, while taking into account the state of the utterance itself and the emotional state of the speaker. Since RNNs cannot achieve parallel computing, Transformers have become a better alternative for sequence modeling \cite{fan2023mgat}. For example, CTNet \cite{lian2021ctnet} uses the powerful representation ability of Transformers to model the emotional dynamics in conversations through a self-attention mechanism. SDT \cite{ma2023transformer} effectively integrates multimodal information through Transformer, and uses self-distillation technology to better learn the potential information in multimodal data.

However, studies have shown \cite{hu2021mmgcn} that the discourse in the conversation is not just a sequential relationship, but a more complex speaker dependency. Therefore, the DialogGCN \cite{ghosal2019dialoguegcn} introduces a graph network to model the dependency between the self and the speaker in the conversation. By using a graph convolutional network (GCN), DialogGCN can effectively propagate contextual information to capture more detailed emotional dependencies. Based on the idea of DialogGCN, SumAggGIN \cite{sheng2020summarize} further emphasizes the emotional fluctuations in the conversation by referencing global topic-related emotional phrases and local dependencies. Meaningwhlie, the DAG-ERC \cite{shen2021directed} believes that the discourse in the conversation is not a simple continuous relationship, but a directed dependency structure.

With the development of pre-trained language models (PLMs) \cite{min2023recent}, researchers began to explore the application of PLM's powerful representation capabilities to emotion recognition tasks. For example, DialogXL \cite{shen2021dialogxl} applies XLNet to emotion recognition and designs an enhanced memory module for storing historical context, while modifying the original self-attention mechanism to capture complex dependencies within and between speakers. The CoMPM \cite{lee2022compm} further leverages PLM by building a pre-trained memory based on the speaker's previous utterances, and then combining the context embedding generated by another PLM to generate the final representation of emotion recognition. CoG-BART \cite{li2022contrast} introduces the BART model to understand the contextual background and generate the next utterance as an auxiliary task.

\subsection{Continuous Graph Neural Networks}

Neural ordinary differential equations (ODEs) are a novel approach to modeling continuous dynamic systems \cite{chen2018neural}. They parameterize the derivatives of hidden states through neural networks, allowing the model to perform continuous inference in the time dimension, rather than relying solely on the discrete sequence of hidden layers in traditional neural networks. ODEs can more accurately describe the changing process over time and are suitable for complex tasks involving time evolution. Continuous Graph Neural Network (CGNN) \cite{xhonneux2020continuous} first extended this ODE approach to graph data. Specifically, CGNN developed a continuous message passing layer to achieve continuous dynamic modeling of node states. Unlike traditional graph neural networks (GCNs), CGNNs no longer rely on a fixed number of layers for information propagation, but instead solve ordinary differential equations to enable continuous propagation of information between nodes. CGNN also introduces a restart distribution to "reset" the node state to the initial state in a timely manner during the information propagation process, thereby avoiding the over-smoothing.

\section{Preliminaries}

\subsection{Graph Neural Networks}

Given a graph $G=(V,E)$, where $V$ is a set of nodes and $E$ is a set of edges. Each node $v \in V$ constitutes the node feature matrix $\mathbf{X}\in \mathbb{R}^{|V|\times d}$, where $d$ represents the dimension of the feature. Each row of $\mathbf{X}$ corresponds to the feature representation of a node. we use the binary adjacency matrix $\mathbf{A}\in \mathbb{R}^{|V|\times |V|}$ to represent the connection relationship between node $i$ and node $j$. If $a_{ij}=1$, it means that there is an edge between node $i$ and node $j$; if $a_{ij}=0$, it means that there is no connection. Our goal is to learn a node representation matrix $\mathbf{H}$ that can capture the structural information and feature information of the nodes in the graph.

We usually normalize the adjacency matrix $\mathbf{A}$. The degree matrix $\mathbf{D}$ is a diagonal matrix whose diagonal elements $\mathbf{D_{ii}}$ represent the degree of node $i$. However, the eigenvalues of the normalized matrix may include negative values. Therefore, we follow the previous methods \cite{kipf2022semi} and use a regularized matrix to represent the graph structure. Specifically, we use the following symmetric normalized adjacency matrix:
\begin{equation}
	\hat{\mathbf{A}}=\frac{\alpha}{2}\left(\mathbf{I}+\mathbf{D}^{-\frac{1}{2}}\mathbf{A}\mathbf{D}^{-\frac{1}{2}}\right)
\end{equation}
where $\alpha$ is a hyperparameter.

\subsection{Neural Ordinary Differential Equation}

Neural ODEs provides a new method for continuous-time dynamic modeling by modeling the forward propagation process of a neural network as the solution process of an ODE. Specifically, consider an input data $x(t)$ and describe its evolution in the form of an ODE:

\begin{equation}
	\frac{dx(t)}{dt}=f(x(t),t,\theta)
\end{equation}
where $x(t)$ represents the hidden state at time $t$, $f$ is a neural function with parameter $\theta$.

\subsection{Multi-modal Feature Extraction}

\textbf{Word Embedding:} Following previous studies \cite{chudasama2022m2fnet, li2022emocaps}, we use RoBERTa \cite{liu2019roberta} to obtain contextual embedding representations of text in this paper.

\textbf{Visual and Audio Feature Extraction:} Following previous work \cite{ma2023transformer, lian2021ctnet}, we selected DenseNet \cite{huang2017densely} and openSMILE \cite{eyben2010opensmile} as feature extraction tools for video and audio. 

\subsection{Problem Definition}
In the multimodal conversational emotion recognition task, given a conversation $C$, the conversation consists of a series of utterances and $S$ different speakers. The goal of multimodal emotion recognition in a conversation is to predict the emotion label of each utterance in the emotion set $Y$. Specifically, the conversation $C$ can be represented as a sequence $C=[(u_1,s_1),(u_2,s_2),\ldots, (u_M,s_M)]$, where $u_i$ represents the $i$-th utterance in the conversation and $s_i$ represents the unique speaker $s_i \in S$ associated with the utterance. Each utterance $u_i$ contains audio data $v_a$, video data $v_f$, and text data $v_t$. These multimodal data together express the meaning and emotion of the utterance. For each utterance $u_i$, we need to determine its emotional state, which is represented by an emotion label $y_i^e$.

\begin{figure*}
	\centering
	\includegraphics[width=1\linewidth]{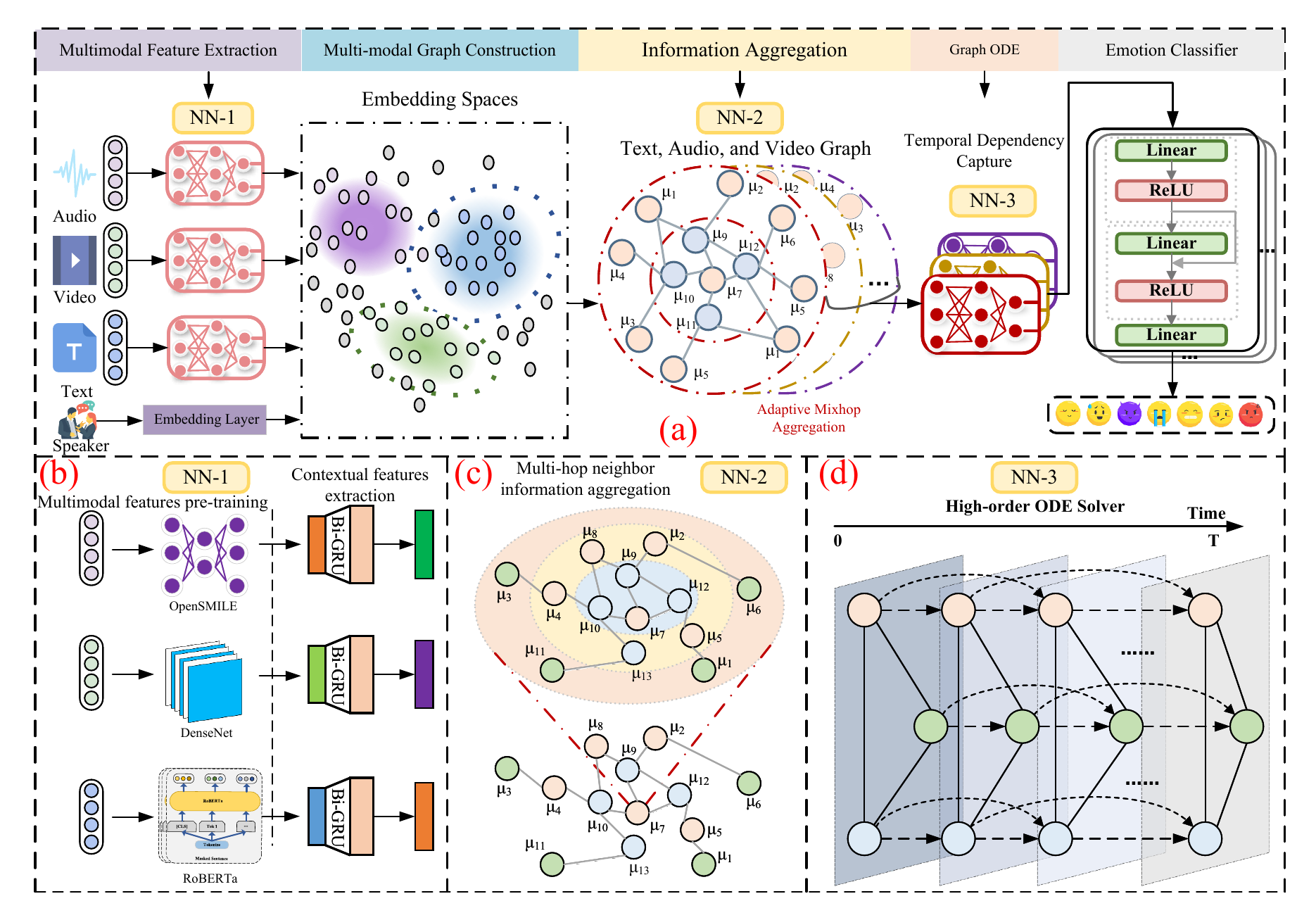}
	\caption{The overall architecture of DGODE.}
	\label{fig:archi}
	\vspace{-3mm}
\end{figure*}

\section{Methodology}
As shown in Fig. \ref{fig:archi}, the key steps of DGODE are to alleviate the overfitting problem of existing GCNs and capture the temporal dependency of speaker emotions. DGODE constructs an adaptive mixhop mechanism in the process of node aggregation to reduce the excessive dependence on local features and thus reduce overfitting. Furthermore, we introduce Graph ODE to model multimodal data in continuous time through differential equations and capture the temporal dependence of speakers. By solving the ODE equation, the emotional state of the previous moment is propagated to the subsequent moment, allowing the model to capture the changing process of the speaker's emotion over a longer time range. In this section, we theoretically introduce the Dynamic Graph Neural Ordinary Differential Equation Network (DGODE) and explain the implementation details.

\subsection{Modality Encoding}

The essence of a conversation is a continuous interactive process in which multiple speakers participate and communicate with each other. Therefore, when processing a conversation, we need to consider the identity of each speaker and the contextual information of the conversation to obtain semantic information that reflects the current discourse and capture the speaker's characteristics and the contextual association information of the conversation. Specifically, we first mark each speaker with a one-hot vector $p_i$ to uniquely identify the speaker. For the $i$-th round of conversation, we extract the corresponding speaker embedding $P_i$ based on the one-hot vector $p_i$, which contains the characteristic information of the current speaker and can be combined with the semantic features of the current discourse to generate a speaker-aware and context-aware unimodal representation. The formula for discourse embedding is defined as follows:
\begin{equation}
	P_i=\mathbf{W}_pp_i
\end{equation}
where $\mathbf{W}_p$ is the learnable parameters.

To effectively encode the features of the conversation text, we use GRU to capture the contextual semantic information in the sequence data and generate a more comprehensive text representation. Specifically, we can make full use of the time order and dependency relationship in the conversation text through GRU, and incorporate the information of the previous and next context into the encoding of each round of conversation, so that the generated text features can better reflect the context and meaning of the entire conversation. Mathematically:
\begin{equation}
	v_m^i=\overleftrightarrow{GRU}(v_m^i,c_{m(+,-)}^i), m \in \{a, v, f\}
\end{equation}
where $c_{m(+,-)}^i$ represents the cell state.

To obtain a unimodal representation that reflects both the speaker identity and the context information, we add the speaker embedding to the representation of each modality. Specifically, for the $i$-th round of speech in the conversation, we calculate the text modality representation $h^i_t$, the audio modality representation $h^i_a$, and the visual modality representation $h^i_f$ and add the speaker embedding $S_i$ to these modality representations to generate the final unimodal representations that incorporate the speaker information as follows:
\begin{equation}
	h^i_m=c^i_m+S_i,\quad m\in\{t,a,v\}
\end{equation}

\subsection{Adaptive MixHop Graph}
The core idea of GCN is to perform convolution operations on graph structured data to capture the complex relationships between nodes and the topological structure of the graph and learn the representation of node features. However, traditional GCN only aggregates information from directly adjacent nodes, and may not be able to fully capture the information of more distant nodes. To capture high-order neighbor relationships, we construct an adaptive mixhop graph to simultaneously extract information from different hop neighbors and improve the understanding of the global graph structure. Furthermore, to model the interaction between different features, we use the residual idea to discretely model the adaptive mixHop GCN as follows:
\begin{equation}
	\mathbf{H}_{n+1}=\sum_{n=1}^{N}\hat{\mathbf{A}}^n\mathbf{H}_n\mathbf{W}+\mathbf{H}_0
	\label{eq3}
\end{equation}
where $\mathbf{W} \in \mathbb{R}^{d \times d}$ represents the learnable weight matrix. Essentially, using the idea of residuals we can model the interaction of different features, so that we can learn the representation of nodes more effectively.

\subsection{Temporal Graph ODE}
However, information aggregation via Eq. \ref{eq3} cannot model the speaker’s emotional changes over time. Therefore, we aim to model the discrete information propagation process of vanilla GCN as a continuous process and use ODE to characterize this dynamic information propagation process. By solving the ODE equation, the emotional state of the previous moment is propagated to the subsequent moments, allowing the model to capture the changing process of the speaker's emotions over a longer time frame. Specifically, we view Eq. \ref{eq3} as the Riemann sum of integrals from $t = 0$ to $t = n$, as described in the following proposition.

\textbf{Proposition 1.}  Suppose $\mathbf{A-I}=\mathbf{P}\mathbf{\Lambda}^{\prime}\mathbf{P}^{-1}$, $\mathbf{W-I}=\mathbf{Q\Phi}^{\prime}\mathbf{Q}^{-1}$, then Eq. \ref{eq3} is discretized as the ODE as follows:
\begin{gather}
	\frac{d\mathbf{H}(t)}{dt}=\frac{1}{N}\sum_{n=1}^{N}\left(ln\hat{\mathbf{A}}\mathbf{H}(t)+\mathbf{H}(t)ln\mathbf{W}+\mathbf{E}\right)\\
	\mathbf{H}(t)=\frac{1}{N}\sum_{n=1}^{N}\left(e^{(\mathbf{A}-\mathbf{I})t}\mathbf{E}e^{(\mathbf{W}-\mathbf{I})t}+\mathbf{P}\mathbf{F}(t)\mathbf{Q}^{-1}\right) 
	\label{eq:ode}
\end{gather}
where $\mathbf{E}=\mathbf{H}(0) = (ln\hat{\mathbf{A}})^{-1}(\hat{\mathbf{A}}-\mathbf{I})\mathbf{E}$, $\mathbf{E} = f(\mathbf{X})$ is the output of the encoder $f$. $\mathbf{F}(t)$ is defined as follows:
\begin{equation}
	\mathbf{F}_{ij}(t)=\frac{\widetilde{\mathbf{E}}_{ij}}{\mathbf{\Lambda}_{ii}^{\mathbf{\prime}}+\mathbf{\Phi}_{jj}^{\mathbf{\prime}}}e^{t(\mathbf{\Lambda}_{ii}^{\prime}+\mathbf{\Phi}_{jj}^{\prime})}-\frac{\widetilde{\mathbf{E}}_{ij}}{\mathbf{\Lambda}_{ii}^{\prime}+\mathbf{\Phi}_{jj}^{\prime}}
\end{equation}
where $\widetilde{\mathbf{E}}=\mathbf{P}^{-1}\mathbf{E}\mathbf{Q}$.

Eq. \ref{eq:ode} can be approximated by an ODE solver to calculate the dynamic evolution of the system in discrete time steps as follows:
\begin{equation}
	\mathbf{H}(t)=\text{ODESolver}(\frac{d\mathbf{H}(t)}{dt},\mathbf{H}_0,t)
\end{equation}

\subsection{Model Training}

The final multi-modal embedding representation $\mathbf{H}_i$ is passed to a fully connected layer for further integration and transformation, and a deeper feature representation is extracted as follows:

\begin{equation}	
	\begin{aligned}
		&\boldsymbol{l}_{i}=\mathrm{ReLU}(\boldsymbol{W}_{l}\mathbf{H}_i+\boldsymbol{b}_{l})\\&\boldsymbol{p}_{i}=\mathrm{softmax}(\boldsymbol{W}_{smax}\boldsymbol{l}_{i}+\boldsymbol{b}_{smax})
	\end{aligned}
\end{equation}
where $\boldsymbol{p}_i$ contains the model's predicted probability for each emotion category, reflecting the model's confidence in identifying different emotions on the utterance, $\boldsymbol{W}_l$, $\boldsymbol{W}_{smax}$, $\boldsymbol{b}_l$, and $\boldsymbol{b}_{smax}$ are trainable parameters. To obtain the final emotion prediction result, we select the emotion category label $\hat{y}_i$ with the highest probability from $\boldsymbol{p}_i$ as the predicted emotion of the utterance as follows:

\begin{equation}
	\hat{y}_i=\arg\max_j(p_{ij})
\end{equation}

\subsection{Implementation Details}
We used PyTorch to implement the proposed DGODE model and chose Adam as the optimizer. For the IEMOCAP dataset, the learning rate of the model was set to 1e-4, while for the MELD dataset, the learning rate was set to 5e-6. During training, the batch size of IEMOCAP was 16, while the batch size of MELD was 8. In the setting of the Bi-GRU layer, we set different numbers of channels for different modal inputs. In the IEMOCAP dataset, the number of input channels for text, acoustic, and visual modalities are 1024, 1582, and 342, respectively. In the MELD dataset, the number of input channels for text, acoustic, and visual modalities are set to 1024, 300, and 342, respectively. In addition, for the graph encoder, we set the size of the hidden layer to 512. To prevent overfitting of the model, we introduced L2 weight decay in training, with the coefficient set to 1e-5, and applied a dropout rate of 0.5 in the key layers.

\begin{table*}[htbp]
	\caption{Comparison with other baselines on the IEMOCAP and MELD dataset.}
	\label{tab:exp}
	\resizebox{\textwidth}{!}{
		\setlength{\tabcolsep}{2pt}{
			\begin{tabular}{lccccccccccccccc}
				\toprule[0.7mm]
				\multirow{2}{*}{Methods} & \multicolumn{7}{c}{IEMOCAP}                                           & \multicolumn{8}{c}{MELD}                                            \\ \cmidrule{2-16} 
				& Happy & Sad  & Neutral & Angry & Excited & Frustrated & W-F1 & Neutral & Surprise & Fear & Sadness & Joy  & Disgust & Anger & W-F1 \\ \midrule[0.4mm]
				bc-LSTM \cite{poria2017context}                   & 34.4  & 60.8 & 51.8    & 56.7  & 57.9    & 58.9       & 54.9 & 73.8    & 47.7     & 5.4  & 25.1    & 51.3 & 5.2     & 38.4  & 55.8 \\
				A-DMN \cite{xing2020adapted}                        & 50.6  & 76.8 & 62.9    & 56.5  & 77.9    & 55.7       & 64.3 & 78.9    & 55.3     & 8.6  & 24.9    & 57.4 & 3.4     & 40.9  & 60.4 \\
				DialogueGCN  \cite{ghosal2019dialoguegcn}              & 42.7  & \textbf{84.5} & 63.5    & 64.1  & 63.1    & 66.9       & 65.6 & 72.1    & 41.7     & 2.8  & 21.8    & 44.2 & 6.7     & 36.5  & 52.8 \\
				RGAT \cite{ishiwatari2020relation}                     & 51.6  & 77.3 & 65.4    & 63.0    & 68.0      & 61.2       & 65.2 & 78.1    & 41.5     & 2.4  & 30.7    & 58.6 & 2.2     & 44.6  & 59.5 \\
				CoMPM \cite{lee2022compm}                           & 60.7  & 82.2 & 63.0      & 59.9  & 78.2    & 59.5       & 67.3 & 82.0      & 49.2     & 2.9  & 32.3    & 61.5 & 2.8     & 45.8  & 63.0 \\
				EmoBERTa \cite{kim2021emoberta}                    & 56.4  & 83.0   & 61.5    & 69.6  & 78.0      & 68.7       & 69.9 & 82.5    & 50.2     & 1.9  & 31.2    & 61.7 & 2.5     & 46.4  & 63.3 \\
				CTNet \cite{lian2021ctnet}                     & 51.3  & 79.9 & 65.8    & 67.2  & \textbf{78.7}    & 58.8       & 67.5 & 77.4    & 50.3     & 10.0   & 32.5    & 56.0   & 11.2    & 44.6  & 60.2 \\
				LR-GCN  \cite{ren2021lr}                   & 55.5  & 79.1 & 63.8    & 69.0    & 74.0      & 68.9       & 69.0 & 80.8    & 57.1     & 0    & 36.9    & \textbf{65.8} & 11.0      & 54.7  & 65.6 \\
				MMGCN  \cite{hu2021mmgcn}                  &  47.1  & 81.9 & 66.4    & 63.5    & 76.2     & 59.1       & 66.8 & 77.0    & 49.6     &  3.6   & 20.4   & 53.8 & 2.8     &  45.2  & 58.4 \\
				AdaGIN  \cite{tu2024adaptive}                     & 53.0    & 81.5 & 71.3    & 65.9  & 76.3    & 67.8       & 70.7 & 79.8    & 60.5     & \textbf{15.2} & 43.7    & 64.5 & \textbf{29.3}    & 56.2  & 66.8 \\
				DER-GCN \cite{ai2024gcn}                  & 58.8  & 79.8 & 61.5    & \textbf{72.1}  & 73.3    & 67.8       & 68.8 & 80.6    & 51.0       & 10.4 & 41.5    & 64.3 & 10.3    & \textbf{57.4}  & 65.5 \\
				M3Net \cite{chen2023multivariate}                   & 60.9  & 78.8 & 70.1    & 68.1  & 77.1    &  67.0       & 71.1 &  79.1    & 59.5       & 13.3     & 42.9 & 65.1    & 21.7 & 53.5 & 65.8\\
				DGODE                 & \textbf{71.8}  & 71.0 & \textbf{74.9}    & 55.7  & 78.6    & \textbf{75.2}       & \textbf{72.8} & \textbf{82.6}    & \textbf{60.9}     & 5.1 & \textbf{45.5}    & 63.4   & 10.6      & 54.0  & \textbf{67.2} \\ \bottomrule[0.7mm]
	\end{tabular}}}
\end{table*}

\section{EXPERIMENTS}
Our experimental results are the average of 10 runs and are statistically significant under paired t-test (all $p < 0.05$).

\subsection{Datasets and Evaluation Metrics}
We used two used MERC datasets in our experiments: IEMOCAP \cite{busso2008iemocap} and MELD \cite{poria2019meld}. Both datasets contain data in three modalities: text, audio, and video. The IEMOCAP dataset is collected from dialogue scenes performed by actors. The MELD dataset consists of dialogue clips from the American TV series Friends. We report the F1 and the weighted F1 (W-F1).

\subsection{Baselines}
To verify the superior performance of our proposed method DGODE, we compared it with other comparison methods, including three RNN algorithms (i.e., bc-LSTM \cite{poria2017context}, A-DMN \cite{xing2020adapted}, CoMPM \cite{lee2022compm}), six GNN algorithms (i.e., DialogueGCN \cite{ghosal2019dialoguegcn}, LR-GCN \cite{ren2021lr}, MMGCN \cite{hu2021mmgcn}, AdaGIN \cite{tu2024adaptive}, DER-GCN \cite{ai2024gcn}, RGAT \cite{ishiwatari2020relation}), one HGNN algorithm (i.e., M3Net \cite{chen2023multivariate}), and two Transformer algorithm (i.e., CT-Net \cite{lian2021ctnet}, EmoBERTa \cite{kim2021emoberta}).

\subsection{Overall Results}

As shown in Table 1, the experimental results show that our proposed method DGODE significantly improves the performance in the emotion recognition task. The performance improvement may be due to the fact that the dynamic graph ODE network can effectively capture the temporal dependency of the discourse and effectively alleviate the over-smoothing problem when processing graph data. To further verify the superiority of the model, we also report the W-F1 of each emotion category. Specifically, in the IEMOCAP dataset, our model has better W-F1 scores than other methods on the three categories of emotions "happy", "neutral" and "frustrated". Similarly, in the MELD dataset, our model also achieves the best W-F1 scores on the three categories of emotions "surprise", "neutral" and "sadness", further verifying the robustness of the model. Therefore, our method not only performs well in emotion recognition tasks, but also has significant advantages in model complexity.

\begin{figure}
	\centering
	\includegraphics[width=1\linewidth]{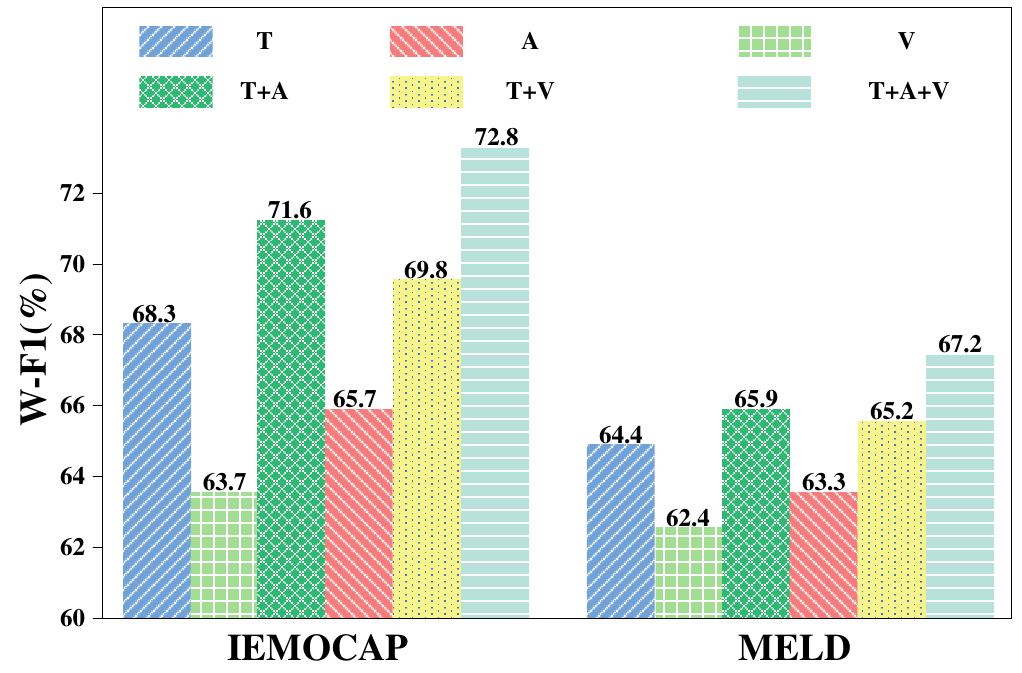}
	\caption{Verify the effectiveness of multimodal features.}
	\label{fig:multimodal}
	\vspace{-4mm}
\end{figure}

\subsection{Effectiveness of Multimodal Features}

We analyze the impact of different modal features on the results of emotion recognition experiments to verify the effect of different modal feature combinations. Specifically, we observe the contribution of different modal features (text, audio, and video) to the emotion recognition performance by inputting them into the model. The experimental results are shown in Fig. \ref{fig:multimodal}. 1) In the single-modal experiment, the emotion recognition accuracy of the text modality is significantly better than the audio and video modalities. 2) When we combine the features of the two modalities for the experiment, the effect of emotion recognition is significantly better than the results of any single modality. 3) When we use the features of the three modalities for emotion recognition at the same time, the performance of the model reaches the best level.

\begin{figure}[htbp]
	\centering
	\subfloat[IEMOCAP]{\includegraphics[width=0.5\linewidth]{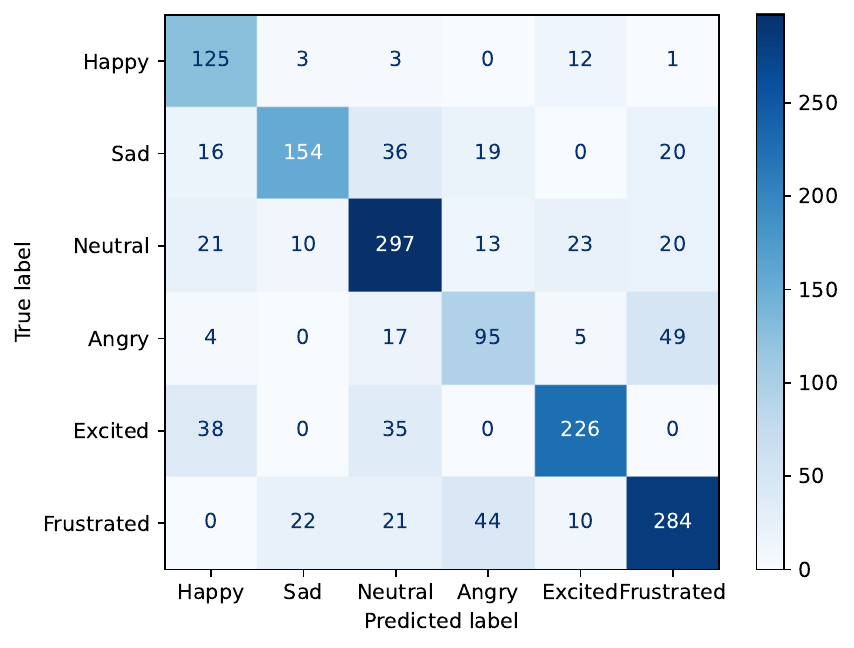}%
		\label{fig:iemo_DGODE}}
	\hfil
	\subfloat[MELD]{\includegraphics[width=0.5\linewidth]{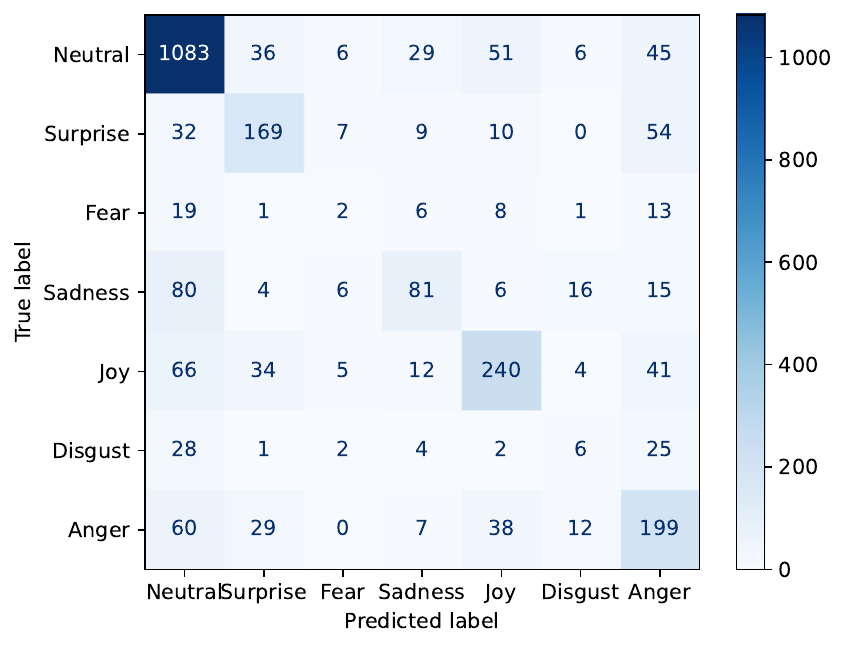}%
		\label{fig:MELD_DGODE}}
	\caption{We performed a analysis of the classification results on the test sets and visualized through confusion matrices.}
	\label{tab:confusion}
	\vspace{-3mm}
\end{figure} 

\begin{figure}[htbp]
	\centering
	\subfloat[IEMOCAP]{\includegraphics[width=0.5\linewidth]{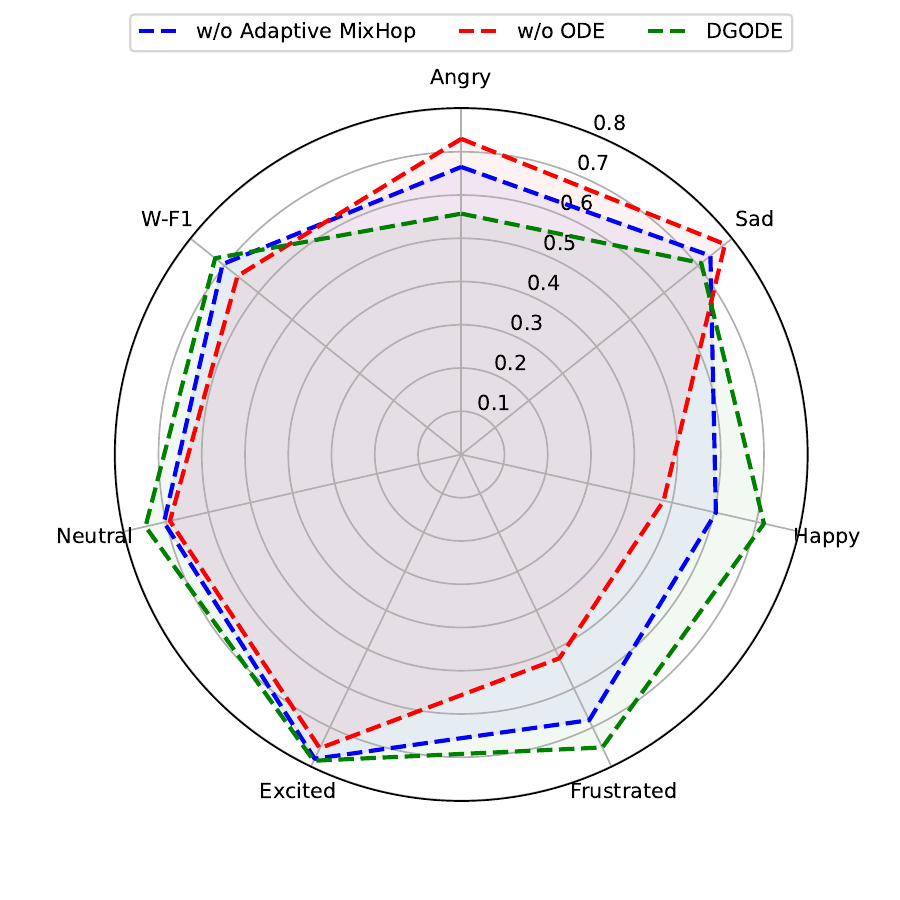}%
		\label{fig:iemo_aba}}
	\hfil
	\subfloat[MELD]{\includegraphics[width=0.5\linewidth]{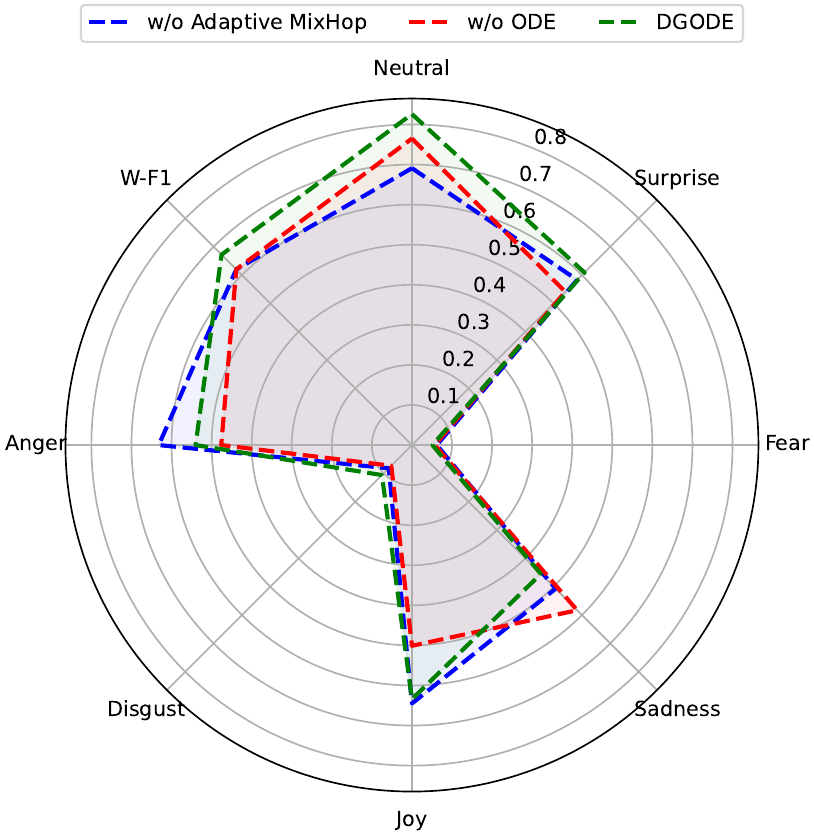}%
		\label{fig:MELD_aba}}
	\caption{On the IEMOCAP and MELD datasets, we performed a detailed analysis of the classification results on the test sets and visualized them through confusion matrices.}
	\label{tab:aba}
	\vspace{-4mm}
\end{figure}

\subsection{Error Analysis}
Although the proposed DGODE model has shown good results in the emotion recognition task, it still faces some challenges, especially in the recognition of some emotions. To analyze the misclassification of the model in more depth, we analyzed the confusion matrix of the test set on the two datasets. As shown in Fig. \ref{tab:confusion}, DGODE has the problem of misclassifying similar emotions on the IEMOCAP dataset. For example, the model often misclassifies "happy" as "excited" or "angry" as "frustrated". The slight differences between emotions lead to the difficulty of the model in distinguishing them. Secondly, on the MELD dataset, DGODE also shows a similar misclassification trend, such as misclassifying "surprise" as "angry". In addition, since the "neutral" emotion is the majority class in the MELD dataset, the model tends to misclassify other emotions as "neutral", which makes the model's performance in dealing with other emotion categories decrease. Finally, the model also encounters significant difficulties in identifying minority emotions. In particular, in the MELD dataset, the two emotions "fear" and "disgust" belong to the minority class, and it is difficult for the model to accurately detect these emotions.

\subsection{Abalation Study}

To analyze the components of DGODE, we performed ablation experiments on the IEMOCAP and MELD datasets. The results in Fig. \ref{tab:aba} show that DGODE consistently outperforms all variants on W-F1 and is also the best on the partially classified sentiment categories. Removing ODE degrades the performance, which highlights the role of ODE in capturing the dynamics of multimodal data. Removing the adaptive mixHop graph also degrades the performance, which emphasizes the importance of capturing high-order relationships.

\begin{figure*}[htbp]
	\centering
	\subfloat[Initial (IEMOCAP)]{\includegraphics[width=0.24\linewidth]{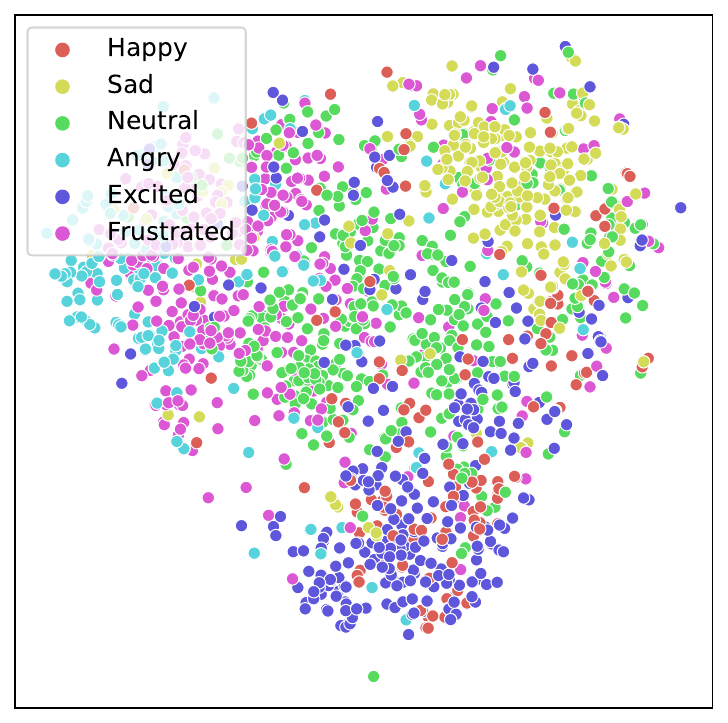}%
		\label{fig:embed_visual_emo_initial_iemocap6}}
	\hfil
	\subfloat[MMGCN (IEMOCAP)]{\includegraphics[width=0.24\linewidth]{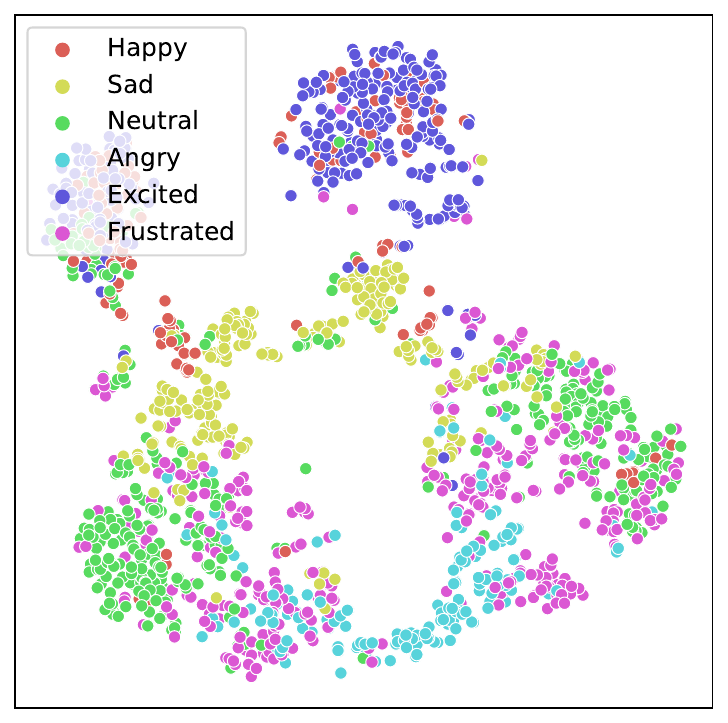}%
		\label{fig:embed_visual_emo_mmgcn_iemocap6}}
	\hfil
	\subfloat[M3Net (IEMOCAP)]{\includegraphics[width=0.24\linewidth]{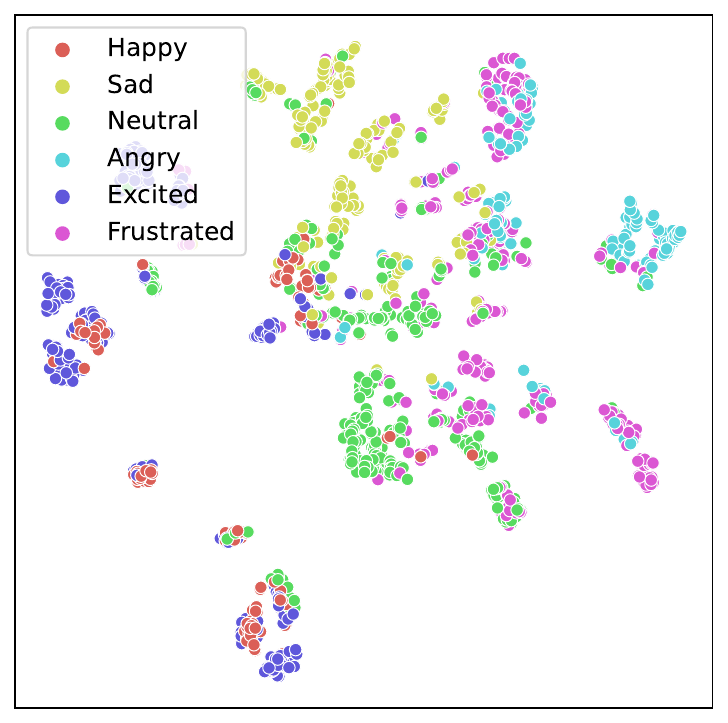}%
		\label{fig:embed_visual_emo_m3net_iemocap6}}
	\hfil
	\subfloat[DGODE (IEMOCAP)]{\includegraphics[width=0.24\linewidth]{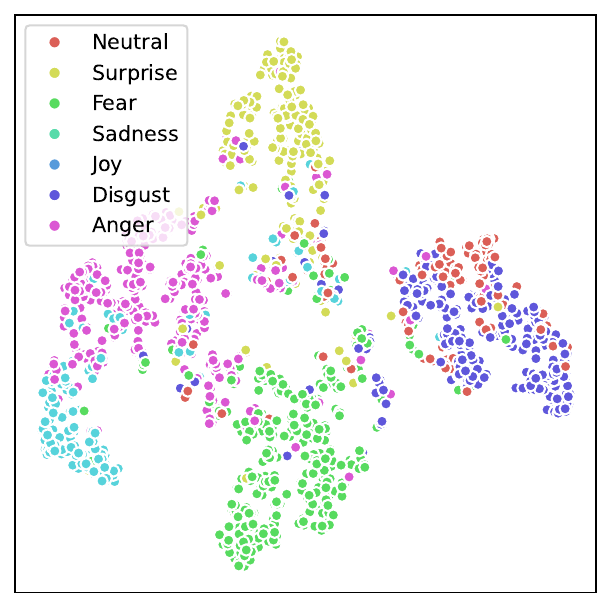}%
		\label{fig:embed_visual_emo_graphsmile_iemocap6}}
	\vfil
	\subfloat[Initial (MELD)]{\includegraphics[width=0.24\linewidth]{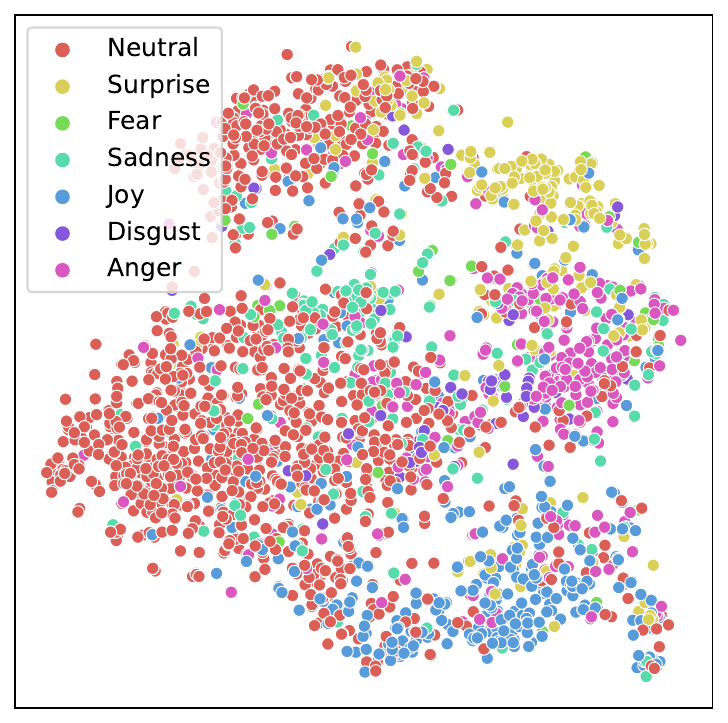}%
		\label{fig:embed_visual_emo_initial_meld}}
	\hfil
	\subfloat[MMGCN (MELD)]{\includegraphics[width=0.24\linewidth]{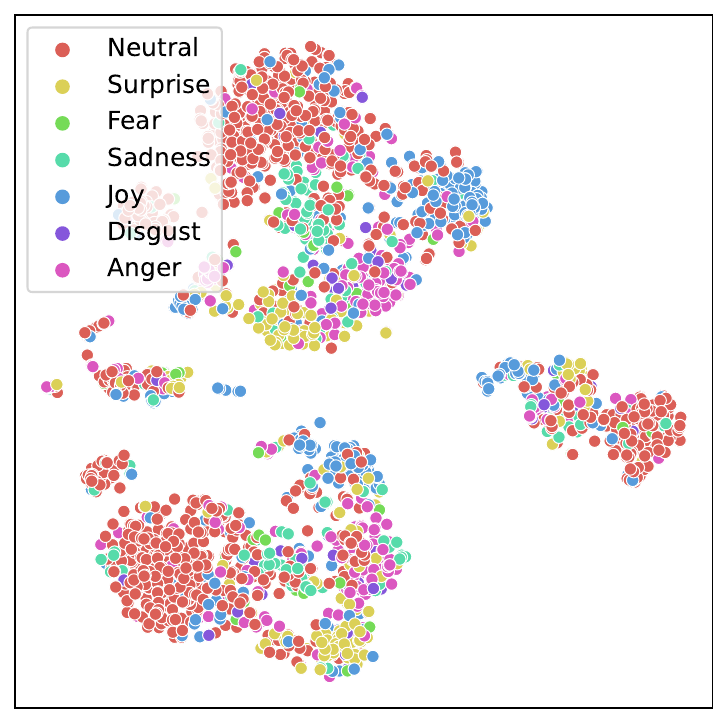}%
		\label{fig:embed_visual_emo_mmgcn_meld}}
	\hfil
	\subfloat[M3Net (MELD)]{\includegraphics[width=0.24\linewidth]{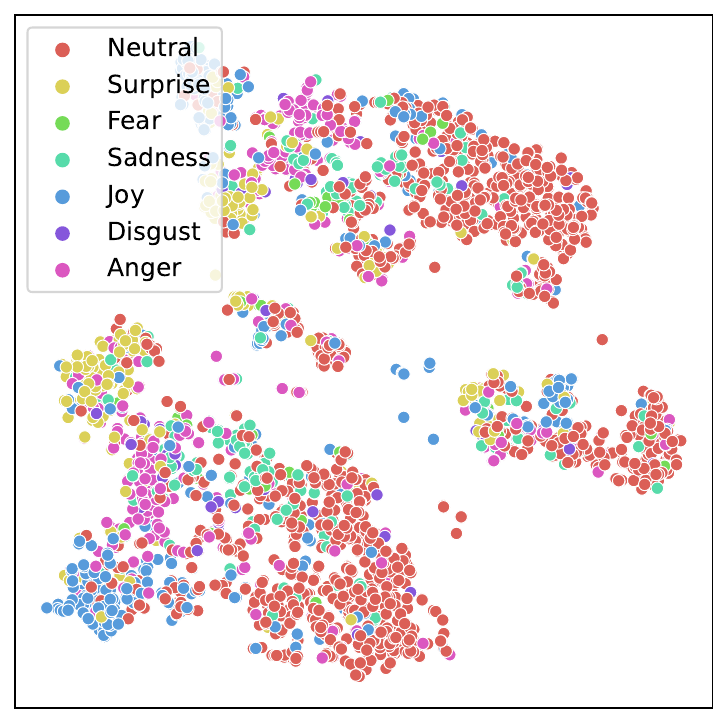}%
		\label{fig:embed_visual_emo_m3net_meld}}
	\hfil
	\subfloat[DGODE (MELD)]{\includegraphics[width=0.24\linewidth]{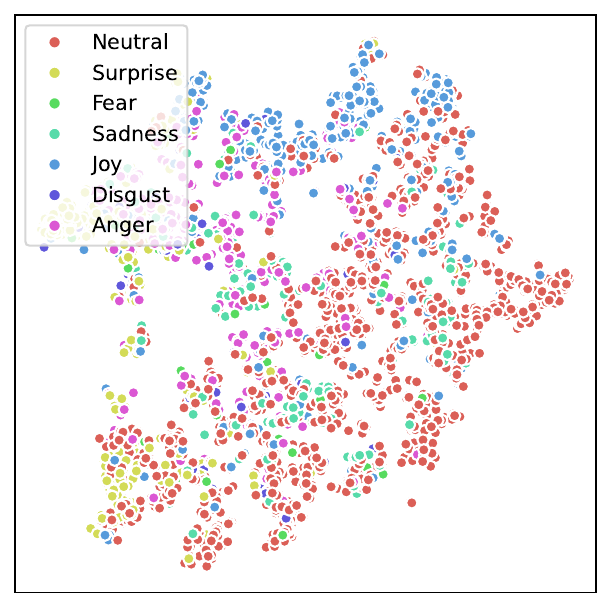}%
		\label{fig:embed_visual_emo_graphsmile_meld}}
	\vfil
	\caption{Visualization of the learned embeddings.}
	\label{fig:embed_visual_emo}
	\vspace{-3mm}
\end{figure*}

\subsection{Visualization}

To more intuitively demonstrate the classification effect of DGODE method in the MERC task, we use T-SNE to visualize the generated sentence vectors. As shown in Fig. \ref{fig:embed_visual_emo}, on the IEMOCAP dataset, the DGODE model performs well, and samples of different emotion categories are effectively separated in the visualization. In contrast, although the MMGCN model can also distinguish samples of different emotion categories to some extent, its classification performance is obviously inferior to DGODE. The distribution of samples generated by MMGCN is relatively chaotic, and the boundaries between different emotion categories are unclear. Meaningwhile, we also compared the classification effect of the M3Net model. Similar to DGODE, M3Net also showed good classification performance on the IEMOCAP dataset, and was able to clearly separate samples of different emotion categories. In the experimental results on the MELD dataset, we observed a similar phenomenon.

\section{Conclusions}

In this paper, we introduce the Dynamic Graph Neural Ordinary Differential Equation Network (DGODE) based on the perspective of controlled diffusion. Our DGODE method introduces an adaptive mixhop mechanism to extract node information from different hop count neighbors simultaneously and uses ordinary differential equations to model the temporal dependence of emotion changes. DGODE shows stable performance as the number of GCN layers increases. We compare DGODE and other baselins on two widely used datasets, and experimental results show that DGODE achieves new SOTA results.

\section{Acknowledgement}
This work is supported by National Natural Science Foundation of China (GrantNo. 69189338), ExcellentYoung Scholars of Hunan Province of China (Grant No. 22B0275), and ChangshaNatural Science Foundation (GrantNo. kq2202294).

\section{Ethical Considerations}

(1) All of our experiments are based on public scientific research datasets that have been widely used in academic research and have undergone strict ethical review. (2) Our research content and experimental design do not involve any sensitive data.

\section{Limitations}

In multimodal emotion recognition, emotion labels are usually annotated for the overall emotion of a certain period of time. However, DGODE focuses on dynamic changes, which may lead to the problem that the subtle dynamic changes captured by the model do not match the overall emotion labels.

\bibliographystyle{coling}
\bibliography{refs}

\clearpage

\appendix

\section{Appendix}
\label{sec:appendix}

\textit{Proof of Proposition 1.} Eq. \ref{eq3} can be rewritten as follows:

\begin{equation}
	\begin{aligned}\mathbf{H}_n&=\sum_{n=1}^{N}\sum_{k=0}^n\mathbf{A}_k^n\mathbf{E}\mathbf{W}_k^n
	\end{aligned}
	\label{eq:7}
\end{equation}

We use Riemann integration to convert Eq. \ref{eq:7} into a continuous form as follows:

\begin{equation}\mathbf{H}(t)=\frac{1}{N}\sum_{n=1}^{N}\int_0^{t+1}\mathbf{A}^s\mathbf{E}\mathbf{W}^s\mathrm{d}s.\end{equation}

Taking the derivative of $\mathbf{H}(t)$ with respect to $t$, we get the following ODE:
\begin{equation}\frac{\mathrm{d}\mathbf{H}(t)}{\mathrm{d}t}=\frac{1}{N}\sum_{n=1}^{N}\mathbf{A}^{t+1}\mathbf{E}\mathbf{W}^{t+1}.\end{equation}

To alleviate the problem of information loss, we take the second-order derivative of $\mathbf{H}(t)$ to obtain an ODE expression with better information aggregation as follows:
\begin{equation}
	\begin{aligned}
		\frac{\mathrm{d}^2\mathbf{H}(t)}{\mathrm{d}t^2}(t)&=\frac{1}{N}\sum_{n=1}^{N}\left(\ln\mathbf{A}\mathbf{A}^{t+1}\mathbf{E}\mathbf{W}^{t+1}\right. \\
		&\left.+\mathbf{A}^{t+1}\mathbf{E}\mathbf{W}^{t+1}\ln\mathbf{W}\right)\\
		&=\frac{1}{N}\sum_{n=1}^{N}\left(\ln\mathbf{A}\frac{\mathrm{d}\mathbf{H}(t)}{\mathrm{d}t}+\frac{\mathrm{d}\mathbf{H}(t)}{\mathrm{d}t}\ln\mathbf{W}\right)
	\end{aligned}
	\label{eq:10}
\end{equation}

Integrating both sides of Eq. \ref{eq:10} with respect to t, we can obtain:
\begin{equation}\frac{\mathrm{d}\mathbf{H}(t)}{\mathrm{d}t}(t)=\ln\mathbf{A}\mathbf{H}(t)+\mathbf{H}(t)\ln\mathbf{W}+c.\end{equation}

The initial value of $\mathbf{H}(0)$ is defined as follows:
\begin{equation}\begin{pmatrix}\mathbf{P}^{-1}\mathbf{H}(0)\mathbf{Q}\end{pmatrix}_{ij}=\frac{\mathbf{\Lambda}_{ii}\widetilde{\mathbf{E}}_{ij}\mathbf{\Phi}_{jj}-\widetilde{\mathbf{E}}_{ij}}{\ln\mathbf{\Lambda}_{ii}+\ln\mathbf{\Phi}_{jj}}
	\label{eq:12}
\end{equation}

When $t=0$, we can get:
\begin{equation}
	\begin{aligned}
		&\frac{\mathrm{d}\mathbf{H}(t)}{\mathrm{d}t}\bigg|_{t=0}=\mathbf{A}\mathbf{E}\mathbf{W} \\
		&\Longrightarrow \mathbf{A}\mathbf{E}\mathbf{W} -\ln\mathbf{A}\mathbf{H}(0)-\mathbf{H}(0)\ln\mathbf{W}=c
	\end{aligned}
	\label{eq:13}
\end{equation}

Combining Eq. 12 and Eq. 13 we can derive:
\begin{equation}
	\begin{aligned}
		\left(\mathbf{P}^{-1}c\mathbf{Q}\right)_{ij}& =\mathbf{\Lambda}_{ii}\widetilde{\mathbf{E}}_{ij}\mathbf{\Phi}_{jj}-\frac{\ln\mathbf{\Lambda}_{ii}(\mathbf{\Lambda}_{ii}\widetilde{\mathbf{E}}_{ij}\mathbf{\Phi}_{jj}-\widetilde{\mathbf{E}}_{ij})}{\ln\mathbf{\Lambda}_{ii}+\ln\mathbf{\Phi}_{jj}} \\
		&-\frac{\mathbf{\Lambda}_{ii}\widetilde{\mathbf{E}}_{ij}\mathbf{\mathbf{\Phi}}_{jj}-\widetilde{\mathbf{E}}_{ij}}{\ln\mathbf{\Lambda}_{ii}+\ln\mathbf{\Phi}_{jj}}\ln\mathbf{\Phi}_{jj} \\
		&c=\mathbf{P}\widetilde{\mathbf{E}}\mathbf{Q}^{-1} =\mathbf{E}
	\end{aligned}
\end{equation}

Therefore, the discrete form of GCN information aggregation can be converted into the continuous form of ODE as follows:
\begin{gather}
	\frac{d\mathbf{H}(t)}{dt}=\frac{1}{N}\sum_{n=1}^{N}\left(ln\hat{\mathbf{A}}\mathbf{H}(t)+\mathbf{H}(t)ln\mathbf{W}+\mathbf{E}\right)
\end{gather}

$\mathbf{H}(t)$ can be further solved by an ODE solver (e.g., the Runge-Kutta method) to obtain.
%
%This is an appendix.

\end{document}